\documentclass{article}
\usepackage{eurospeech_01,amssymb,amsmath,graphicx}
\ninept
\def\reg{{\rm\ooalign{\hfil
     \raise.07ex\hbox{\scriptsize R}\hfil\crcr\mathhexbox20D}}}

\title{Building a Test Collection for Speech-Driven Web Retrieval}

\makeatletter
\def\name#1{\gdef\@name{#1\\}}
\makeatother
\name{{\em Atsushi Fujii$^{\dagger,\dagger\dagger\dagger}$, Katunobu Itou$^{\dagger\dagger,\dagger\dagger\dagger}$}}

\address{$^{\dagger}$ Institute of Library and Information Science,
  University of Tsukuba \\
  $^{\dagger\dagger}$ National Institute of Advanced Industrial Science
  and Technology \\
  $^{\dagger\dagger\dagger}$ CREST, Japan Science and Technology
  Corporation}
\newcommand{\eq}[1]{(\ref{#1})}

\begin{document}
\maketitle
\begin{abstract}
  This paper describes a test collection (benchmark data) for
  retrieval systems driven by spoken queries. This collection was
  produced in the subtask of the NTCIR-3 Web retrieval task, which was
  performed in a TREC-style evaluation workshop. The search topics and
  document collection for the Web retrieval task were used to produce
  spoken queries and language models for speech recognition,
  respectively. We used this collection to evaluate the performance of
  our retrieval system.  Experimental results showed that (a) the use
  of target documents for language modeling and (b) enhancement of the
  vocabulary size in speech recognition were effective in improving
  the system performance.
\end{abstract}

\section{Introduction}
\label{sec:introduction}

Automatic speech recognition, which decodes the human voice to
generate transcriptions, has recently become a practical technology.
A number of speech-based methods have been explored in the information
retrieval (IR) community, which can be classified into the following
two fundamental categories:
\begin{itemize}
\item spoken document retrieval, in which written queries are used to
  search speech (e.g., broadcast news audio) archives for relevant
  speech information,
\item speech-driven retrieval, in which spoken queries are used to
  retrieve relevant textual
  information.
\end{itemize}
Initiated partially by the TREC-6 spoken document retrieval (SDR)
track~\cite{garofolo:trec-97}, various methods have been proposed for
spoken document retrieval.  However, a relatively small number of
methods~\cite{barnett:eurospeech-97,crestani:fqas-2000,fujii:emnlp-2002}
have been explored for speech-driven text retrieval, although they are
associated with numerous keyboard-less retrieval applications, such as
telephone-based retrieval, car navigation systems, and user-friendly
interfaces.

In the NTCIR-3
workshop\footnote{http://research.nii.ac.jp/ntcir/index-en.html},
which is a TREC-style evaluation workshop, the Web retrieval main task
was organized to promote text-based Web
IR~\cite{eguchi:sigir-2002}. Additionally, {\em optional\/} subtasks
were also invited, in which a group of researchers voluntarily
organized a subtask to promote their common research area.  We made
use of this opportunity and organized the ``speech-driven retrieval''
subtask to produce a reusable test collection for experimental of Web
retrieval driven by spoken queries.

Section~\ref{sec:test_collection} describes the test collection
produced for the speech-driven retrieval subtask.
Section~\ref{sec:system} describes our speech-driven retrieval system,
and Section~\ref{sec:experimentation} elaborates on comparative
experiments, in which we evaluated our system in terms of the speech
recognition and retrieval accuracy.

\section{Test Collection for Speech-Driven IR}
\label{sec:test_collection}

\subsection{Overview}
\label{subsec:test_collection_overview}

The purpose of the speech-driven retrieval subtask was to produce
reusable and publicly available test collections and tools, so that
researchers in the information retrieval and speech processing
communities can develop technologies and share scientific knowledge
concerning speech-driven information retrieval.  In principle, as with
conventional IR test collections, test collections for speech-driven
retrieval are required to include test queries, target documents, and
relevance assessment for each query.  However, unlike conventional
text-based IR, queries are speech data uttered by humans.  In
practice, because producing the entire collection is prohibitive, we
produced speech data related to the Web retrieval main (text-based)
task. Thus, target documents and relevance assessment in the main task
can be used for the purpose of speech-driven retrieval.

However, participants for the NTCIR workshop are mainly researchers in
the information retrieval and natural language processing communities,
and are not necessarily experts in developing and operating speech
recognition systems. Therefore, we also produced language models that
can be used with an existing speech recognition engine (decoder),
which helps researchers to perform experiments similar to those
described in this paper.  All above data are included in the NTCIR-3
Web retrieval test collection, which is publicly available.

\subsection{Spoken Queries}
\label{subsec:spoken_queries}

For the Web retrieval main task, 105 search topics were produced, for
each of which relevance assessment was performed with respect to two
different document sets: the 10GB and 100GB collections. The 10GB and
100GB collections correspond approximately to 1M and 10M documents,
respectively.

Each topic is in SGML-style form and consists of the topic ID
(\verb|<NUM>|), title of the topic (\verb|<TITLE>|), description
(\verb|<DESC>|), narrative (\verb|<NARR>|), list of synonyms related
to the topic (\verb|<CONC>|), sample of relevant documents
(\verb|<RDOC>|), and a brief profile of the user who produced the topic
(\verb|<USER>|).  Figure~\ref{fig:topic} depicts a translation of an
example topic. Although Japanese topics were used in the main task,
English translations are also included in the Web retrieval collection
mainly for publication purposes.

\begin{figure}[htbp]
  \begin{center}
    \leavevmode
    \scriptsize
    \begin{quote}
      \tt
      <TOPIC> \\
      <NUM>0010</NUM> \\
      <TITLE CASE="b">Aurora, conditions, observation</TITLE> \\
      <DESC>For observation purposes, I want to know the conditions
      that give rise to an aurora</DESC> \\
      <NARR><BACK>I want to observe an aurora so I want to know the
      conditions necessary for its occurrence and the mechanism behind
      it.</BACK><RELE>Aurora observation records, etc. list the place
      and time so only documents that provide additional information
      such as the weather and temperature at the time of occurrence
      are relevant. </RELE></NARR> \\
      <CONC>Aurora, occurrence, conditions, observation,
      mechanism</CONC> \\
      <RDOC>NW003201843, NW001129327, NW002699585</RDOC> \\
      <USER>1st year Master's student, female, 2.5 years search
      experience</USER> \\
      </TOPIC>
    \end{quote}
    \caption{An example topic in the Web retrieval collection.}
    \label{fig:topic}
  \end{center}
\end{figure}

Participants in the main task were allowed to submit more than one
retrieval result using one or more fields. However, participants were
required to submit results obtained with the title and description
fields independently. Titles are lists of keywords, and descriptions
are phrases and sentences.

From the viewpoint of speech recognition, titles and descriptions can
be used to evaluate {\em word\/} and {\em continuous\/} recognition
methods, respectively. Because state-of-the-art speech recognition is
based on a continuous recognition framework, we used only the
description field.  For the first speech-driven retrieval subtask, we
focused on {\em dictated\/} ({\em read\/}) speech, although our
ultimate goal is to recognize {\em spontaneous\/} speech. We asked ten
speakers (five adult males and five adult females) to dictate
descriptions in the 105 topics.  The ten speakers also dictated 50
sentences in the ATR phonetic-balanced sentence set as reference data,
which can potentially be used for speaker adaptation. However, we did
not use this additional data for the purpose of the experiments
described in this paper.  The above-mentioned spoken queries and
sentences were recorded with the same close-talk microphone in a
noiseless office. Speech waves were digitized at a 16KHz sampling
frequency and a quantization of 16 bits. The resulting data are in the
RIFF format.

\subsection{Language Models}
\label{subsec:lang_model}

Unlike general-purpose speech recognition, in speech-driven text
retrieval, users usually speak contents associated with a target
collection, from which documents relevant to user needs are retrieved.
In a stochastic speech recognition framework, the accuracy depends
primarily on acoustic and language models.  Whereas acoustic models
are related to phonetic properties, language models, which represent
linguistic contents to be spoken, are related to target
collections. Therefore, it is feasible that language models have to be
produced based on target collections. In summary, our belief is that
by adapting a language model to a target IR collection, we can improve
the speech recognition accuracy and, consequently, the retrieval
accuracy.  Motivated by this background, we used target documents for
the main task to produce the language models. For this purpose, we
used only the 100GB collection, because the 10GB collection is a
subset of the 100GB collection.

We produced two language models of different vocabulary sizes so that
the relation between the vocabulary size and system performance can be
investigated. In practice, 20K and 60K high frequency words were used
independently to produce word-based trigram models. We shall call
these models ``Web20K'' and ``Web60K'', respectively. We used the
ChaSen morphological analyzer\footnote{http://chasen.aist-nara.ac.jp/}
to extract words from the 100GB collection.  To resolve the data
sparseness problem, we used a back-off smoothing method, in which the
Witten-Bell discounting method was used to compute back-off
coefficients. In addition, through preliminary experiments, cut-off
thresholds were empirically set at 20 and 10 for the Web20K and Web60K
models, respectively. Trigrams whose frequency was above the threshold
were used for language modeling. Language models and dictionaries are
in the ARPA and HTK formats, respectively.

Table~\ref{tab:lang_model} shows the statistics related to word
tokens/types in the 100GB collection and ten years of ``Mainichi
Shimbun'' newspaper articles from 1991 to 2000. We shall use the term
``word token'' to refer to occurrences of words, and the term ``word
type'' to refer to vocabulary items. The size of the 100G collection
(``Web'') is approximately 10 times that of 10 years of newspaper
articles (``News''), which was one of the largest Japanese corpora
available for the purpose of research and development in language
modeling. This means that the Web is a vital, as yet untapped, corpus
for language modeling.

\begin{table}[htbp]
  \begin{center}
    \caption{The statistics of corpora for language modeling.}
    \medskip
    \leavevmode
    \small
    \begin{tabular}{lccc} \hline\hline
      & Web (100GB) & News (10 years) \\ \hline
      \# of Word types & 2.57M & 0.32M \\
      \# of Word tokens & 2.44G & 0.26G \\
      \hline
    \end{tabular}
    \label{tab:lang_model}
  \end{center}
\end{table}

\section{System Description}
\label{sec:system}

\subsection{Overview}
\label{subsec:system_overview}

\begin{figure}[htbp]
  \begin{center}
  \leavevmode
  \includegraphics[height=1.9in]{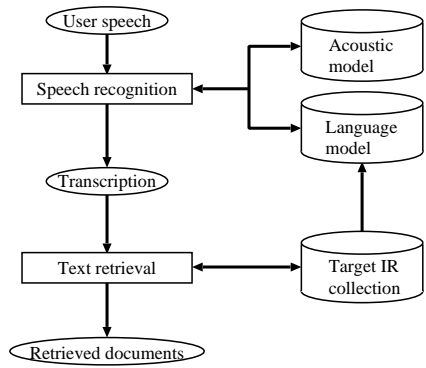}
  \end{center}
  \caption{An overview of our speech-driven retrieval system.}
  \label{fig:system}
\end{figure}

Figure~\ref{fig:system} depicts the overall design of our
speech-driven text retrieval system, which consists of speech
recognition and text retrieval modules.  In the off-line process, a
target IR collection is used to produce a language model, so that user
speech related to the collection can be recognized with high
accuracy. However, an acoustic model was produced independently of the
target collection.  In the on-line process, given an information
request spoken by a user (i.e., a spoken query), the speech
recognition module uses acoustic and language models to generate a
transcription of the user speech. Then, the text retrieval module
searches the target IR collection for documents relevant to the
transcription, and outputs a specific number of top-ranked documents
according to the degree of relevance in descending order.  In the
following two sections, we describe the speech recognition and text
retrieval modules.

\subsection{Speech Recognition}
\label{subsec:speech_recognition}

We used the Japanese dictation
toolkit\footnote{http://winnie.kuis.kyoto-u.ac.jp/dictation/}
including the Julius decoder and acoustic/language models. Julius
performs a two-pass (forward-backward) search using word-based forward
bigrams and backward trigrams. The acoustic model was produced from
the ASJ speech database, which contains 20,000 sentences uttered by
132 speakers including both genders. A 16-mixture Gaussian
distribution triphone Hidden Markov Model, in which the states are
clustered into 2,000 groups by a state-tying method, is used.  The
language model is a word-based trigram model produced from 60,000 high
frequency words in 10 years of Mainichi Shimbun newspaper articles.
This toolkit also includes development software so that acoustic and
language models can be produced depending on the application.  While
we used the acoustic model provided in the toolkit, we used new
language models produced from the 100GB collections, that is, the
Web20K and Web60K models.

\subsection{Text Retrieval}
\label{subsec:text_retrieval}

The retrieval module is based on an existing retrieval
method~\cite{robertson:sigir-94}, which computes the relevance score
between the transcribed query and each document in the collection.
The relevance score for document $d$ is computed by
Equation~\eq{eq:okapi}.
\begin{equation}
  \footnotesize
  \label{eq:okapi}
  \sum_{t} f_{t,q}\cdot\frac{\textstyle (K+1)\cdot
  f_{t,d}}{K\cdot\{(1-b)+\textstyle\frac{\textstyle
  dl_{d}}{\textstyle b\cdot avgdl}\} +
  f_{t,d}}\cdot\log\frac{\textstyle N - n_{t} + 0.5}{\textstyle
  n_{t} + 0.5}
\end{equation}
where $f_{t,q}$ and $f_{t,d}$ denote the frequency that term $t$
appears in query $q$ and document $d$, respectively; $N$ and $n_{t}$
denote the total number of documents in the collection and the number of
documents containing term $t$, respectively; $dl_{d}$ denotes the
length of document $d$, and $avgdl$ denotes the average length of
documents in the collection. We empirically set $K=2.0$ and $b=0.8$,
respectively.

Given transcriptions (i.e., speech recognition results for spoken
queries), the retrieval module searches a target IR collection for
relevant documents and sorts them in descending order according to the
score.  We used content words, such as nouns, extracted from documents
as index terms, and performed word-based indexing.  We used the ChaSen
morphological analyzer to extract content words. We also extracted
terms from transcribed queries using the same method. We used words
and bi-words (i.e., word-based bigrams) as index terms.

\section{Experimentation}
\label{sec:experimentation}

\begin{table*}[htbp]
  \begin{center}
    \caption{Experimental results for different retrieval methods
      targeting the 10GB and 100GB collections (OOV: test-set
    out-of-vocabulary rate, WER: word error rate, TER: term error
    rate, MAP: mean average precision).}
    \medskip
    \leavevmode
    \small
    \tabcolsep=5pt
    \begin{tabular}{lccccccccccc} \hline\hline
      & & & & \multicolumn{4}{c}{MAP (10GB)}
      & \multicolumn{4}{c}{MAP (100GB)} \\ \cline{5-12}
      {\hfill\centering Method\hfill}
      & OOV & WER & TER
      & RC & RL & PC & PL
      & RC & RL & PC & PL \\ \hline
      Text & --- & --- & --- & .1470 & .1286 & .1612 & .1476 
      & .0855 & .0982 & .1257 & .1274 \\
      Web60K & .0073 & .1311 & .2162 &
      .0966 & .0916 & .0973 & .1013 & .0542 & .0628 & .0766 & .0809 \\
      News60K & .0157 & .1806 & .2991 &
      .0701 & .0681 & .0790 & .0779 & .0341 & .0404 & .0503 & .0535 \\
      Web20K & .0423 & .1642 & .2757 &
      .0616 & .0628 & .0571 & .0653 & .0315 & .0378 & .0456 & .0485 \\
      \hline
    \end{tabular}
    \label{tab:results}
  \end{center}
\end{table*}

In the Web retrieval main task, different types of text retrieval were
performed. The first type was ``Topic Retrieval'' resembling the TREC
ad hoc retrieval. The second type was ``Similarity Retrieval'', in
which documents were used as queries instead of keywords and phrases.
The third type was ``Target Retrieval'', in which systems with a high
precision were highly valued. This feature provided a salient contrast
to the first two retrieval types, in which both recall and precision
were used equally as evaluation measures.

Although the spoken queries produced can be used for the first and
third task types, we focused solely on Topic Retrieval for the sake of
simplicity. We used the 47 topics for the Topic Retrieval task to
retrieve the 1,000 top documents, and we used the TREC evaluation
software to calculate the mean average precision (MAP) values (i.e.,
non-interpolated average precision values, averaged over the 47
topics).

Relevance assessment was performed based on four ranks of relevance:
highly relevant, relevant, partially relevant and irrelevant.  In
addition, unlike conventional retrieval tasks, documents hyperlinked
from retrieved documents were optionally used for relevance
assessment. In summary, the following four assessment types were
available to calculate the MAP values:
\begin{itemize}
\item (highly) relevant documents were regarded as correct answers,
  and hyperlink information was not used (RC),
\item (highly) relevant documents were regarded as correct answers,
  and hyperlink information was used (RL),
\item partially relevant documents were also regarded as correct
  answers, and hyperlink information was not used (PC),
\item partially relevant documents were also regarded as correct
  answers, and hyperlink information was used (PL).
\end{itemize}
In the formal run for the main task, we submitted results obtained
with different methods for the 10GB and 100GB collections. The best
performance was obtained when we used description (\verb|<DESC>|)
fields as queries and we used a combination of words and bi-words as
index terms.

The purpose of the experiments for speech-driven retrieval was
two-fold. First, we investigated the extent to which a language model
based on a target document collection contributes to an improvement in
performance. Second, we investigated the impact of the vocabulary size
for speech recognition on speech-driven retrieval. Therefore, we
compared the performance of the following four retrieval methods:
\begin{itemize}
\item text-to-text retrieval, which used written queries, and can be
  seen as the perfect speech-driven text retrieval method (``Text''),
\item speech-driven text retrieval, in which the Web60K model was used
  (``Web60K''),
\item speech-driven text retrieval, in which a language model produced
  from 60,000 high frequency words in ten years of Mainichi Shimbun
  newspaper articles was used (``News60K''),
\item speech-driven text retrieval, in which the Web20K model was used
  (``Web20K'').
\end{itemize}
For text-to-text retrieval, we used descriptions (\verb|<DESC>|) as
queries, because the spoken queries used for speech-driven retrieval
methods were descriptions dictated by speakers.

For speech-driven text retrieval methods, queries dictated by the ten
speakers were used independently, and the final result was obtained by
averaging the results for all speakers.  Although the Julius decoder
used in the speech recognition module generated more than one
transcription candidate (hypothesis) for a single speech, we used only
that with the greatest probability score.  All language models were
produced by means of the same softwares, but they were different in
terms of the vocabulary size and the source documents.
Table~\ref{tab:results} shows the MAP values with respect to the four
relevance assessment types and the word error rate in speech
recognition, for different retrieval methods targeting the 10GB and
100GB collections.

As with existing experiments for speech recognition, the word error
rate (WER) is the ratio between the number of word errors (i.e.,
deletion, insertion, and substitution) and the total number of
words. In addition, we investigated the error rate with respect to
query terms (i.e., keywords used for retrieval), which we shall call
the term error rate (TER). Note that unlike MAP, smaller values of WER
and TER are obtained with better methods.  Table~\ref{tab:results}
also shows the test-set out-of-vocabulary rate (OOV), which is the
ratio of the number of words not included in the speech recognition
dictionary to the total number of words in the spoken queries.
Suggestions that can be derived from the results in
Table~\ref{tab:results} are as follows.

Looking at the WER and TER columns, News60K and Web20K were comparable
in speech recognition performance, but Web60K outperformed in both
cases. However, the difference between News60K and Web20K in OOV did
not affect WER and TER. In addition, TER was greater than WER, because
in computing TER, functional words, which are generally recognized
with a high accuracy, were excluded.

Whereas the MAP values of News60K and Web20K were comparable, the MAP
values of Web60K, which were approximately 60--70\% of those obtained
with Text, were greater than those for News60K and Web20K,
irrespective of the relevance assessment type.  These results were
observed for both the 10GB and 100GB collections.

The only difference between News60K and Web60K was the source corpus
for language modeling in speech recognition, and therefore we conclude
that the use of target collections to produce a language model was
effective for speech-driven retrieval. In addition, by comparing the
MAP values of Web20K and Web60K, we conclude that the vocabulary size
for speech recognition was also influential for the performance of
speech-driven retrieval.

We analyzed speech recognition errors, focusing mainly on those
attributed to the out-of-vocabulary problem. Table~\ref{tab:oov} shows
the ratio of the number of out-of-vocabulary words to the total number
of misrecognized words (or terms) in transcriptions. However, it
should be noted that the actual ratio of errors due to the OOV problem
can potentially be higher than those figures, because non-OOV words
collocating with OOV words are often misrecognized. The remaining
reasons for speech recognition errors are associated with insufficient
N-gram statistics and the acoustic model.  As predicted, the ratio of
OOV words (terms) in Web20K was much higher than the ratios in Web60K
and News60K.  However, by comparing News60K and Web20K, WER and TER of
News60K in Table~\ref{tab:results} were higher than those of
Web20K. This suggests that insufficient N-gram statistics were more
problematic in News60K, compared to Web20K.

\begin{table}[htbp]
  \begin{center}
    \caption{The ratio of the number of OOV words/terms to the total
      number of misrecognized words/terms.}
    \medskip
    \leavevmode
    \small
    \tabcolsep=5pt
    \begin{tabular}{lcc} \hline\hline
      & Word & Term \\ \hline
      Web60K & .0704 & .1838 \\
      News60K & .0966 & .2143 \\
      Web20K & .2855 & .5049 \\
      \hline
    \end{tabular}
    \label{tab:oov}
  \end{center}
\end{table}

\section{Conclusion}
\label{sec:conclusion}

In the NTCIR-3 Web retrieval task, we organized the speech-driven
retrieval subtask and produced 105 spoken queries dictated by ten
speakers. We also produced word-based trigram language models using
approximately 10M documents in the 100GB collection used for the main
task.  We used those queries and language models to evaluate the
performance of our speech-driven retrieval system.  Experimental
results showed that (a) the use of target documents for language
modeling and (b) enhancement of the vocabulary size in speech
recognition were effective in improving the system performance. Future
work will include experiments using spontaneous spoken queries.

\section{Acknowledgments}

The authors thank the organizers of the NTCIR-3 Web retrieval task for
their support to the speech-driven retrieval subtask.

\bibliographystyle{IEEEtran}

\end{document}